\documentclass[runningheads]{llncs}

\usepackage{eccv}

\usepackage{eccvabbrv}

\usepackage{graphicx}
\usepackage{booktabs}

\usepackage[accsupp]{axessibility}  

\usepackage{hyperref}

\usepackage{orcidlink}

\usepackage{subcaption} 
\usepackage{booktabs, multirow, colortbl, xcolor}
\usepackage{wrapfig}

\begin{document}

\title{FocusVLA: Focused Visual Utilization for Vision-Language-Action Models} 

\titlerunning{FocusVLA}

\author{
Yichi Zhang \inst{1,2} * \and
Weihao Yuan \inst{2,3} *$^\dagger$  \and
Yizhuo Zhang \inst{4} \and
Xidong Zhang\inst{1,2} \and
Jia Wan\inst{1}
}

\authorrunning{Y. Zhang et al.}

\institute{
Harbin Institute of Technology, Shenzhen, China \and
DaiMon Robotics, China \and
Nanjing University, Nanjing, China \and
Renmin University of China, Beijing, China
}


\maketitle

\begin{abstract}
Vision-Language-Action (VLA) models improve action generation by conditioning policies on rich vision-language information. However, current auto-regressive policies are constrained by three bottlenecks:
(1) architectural bias drives models to overlook visual details,
(2) an excessive number of visual tokens makes attention difficult to focus on the correct regions,
and (3) task-irrelevant visual information introduces substantial noise -- together severely impairing the quality of action.
In this paper, we investigate how to effectively utilize different visual representations for action generation. To this end, we first empirically validate the above issues and show that VLA performance is primarily limited by how visual information is utilized, rather than by the quality of visual representations.
Based on these insights, we introduce FocusVLA, a novel paradigm that directs the model’s attention to task-relevant visual regions to effectively bridge vision to action. 
Specifically, we first propose Modality Cascaded Attention to eliminate shortcut pathways, thereby compelling VLA models to rely on task-relevant visual details for action generation.
Furthermore, we propose Focus Attention, which dynamically selects task-relevant visual patches to control information quantity while explicitly modulating their influence to suppress task-irrelevant noise.
Extensive experiments on both simulated and real-world robotic benchmarks demonstrate that FocusVLA not only effectively leverages visual details to perform dexterous manipulations, but also substantially improves performance and accelerates convergence across a variety of tasks.
The code of our method will be made public.
\keywords{Vision-Language-Action \and Task-Relevant Visual Utilization}
\end{abstract}

\let\thefootnote\relax\footnotetext{* Equal Contribution}
\let\thefootnote\relax\footnotetext{$\dagger$ Project Lead}

\begin{figure}[t]
  \centering
  \includegraphics[width=1.\linewidth]{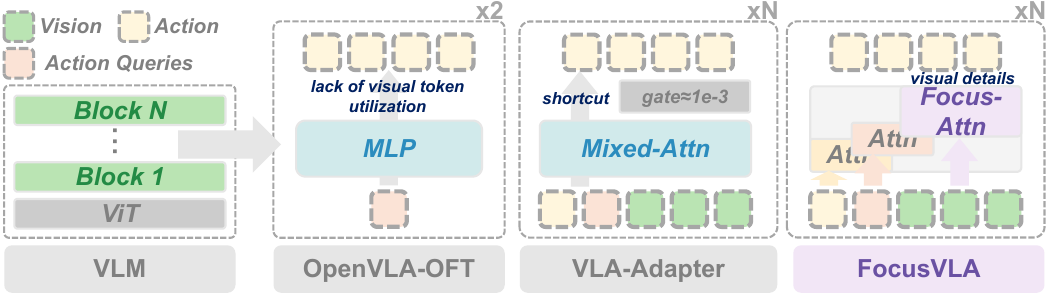}
  \vspace{-4mm}
  \caption{Illustration of structural limitations in existing auto-regressive Vision-Language-Action (VLA) models and the motivation of FocusVLA. OpenVLA-OFT suffers from the lack of direct utilization of visual tokens, limiting its ability to ground actions in fine-grained visual details. VLA-Adapter's mixed attention introduces architectural shortcuts that enable the model to bypass concrete visual details by favoring the easier action-query pathway, leading to imprecise manipulation. Furthermore, it introduces a near-zero gating factor that substantially suppresses visual signals.
  In contrast, FocusVLA uses Modality Cascaded Attention to enforce sequential modality interaction, ensuring the model relies on task-relevant visual details before action reasoning and eliminating shortcut pathways. In addition, we propose Focus Attention to suppress task-irrelevant information, leading to precise and robust manipulation.}
  \vspace{-4mm}
  \label{fig:model_comparison}
\end{figure}

\begin{figure}[t]
  \centering
  \includegraphics[width=1.\linewidth]{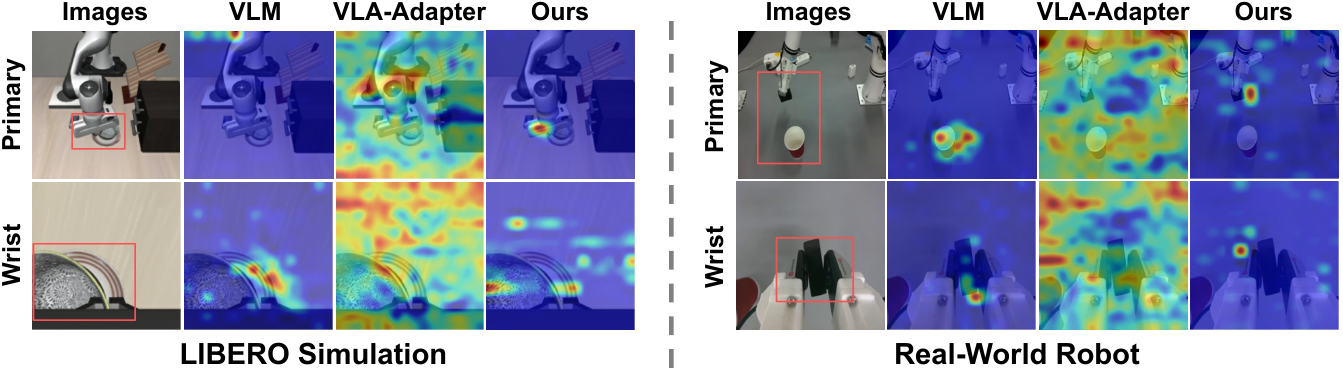}
  \vspace{-4mm}
  \caption{Visualization of attention maps on LIBERO simulation (left) and real-world environments (right). We extract the attention scores from the last layer, average them across attention heads and queries, and project them back onto the image plane for visualization. Specifically, for VLM, we use the attention from the action query to visual tokens, while for the policy, we use the attention from the action latent to visual tokens. VLA-Adapter exhibits highly scattered and distracted attention patterns, largely focusing on task-irrelevant regions due to structural shortcuts, thereby failing to capture concrete visual details. In contrast, FocusVLA produces concentrated and task-aligned attention maps that consistently focus on contact regions and manipulation targets. This targeted attention enables more precise action generation across both simulated and real-world settings.}
  \vspace{-3mm}
  \label{fig:vis_attn_map}
\end{figure}

\vspace{-6mm}
\section{Introduction}
\label{sec:intro}
Vision-Language-Action (VLA) models have become a dominant paradigm for robotic manipulation, enabling policies to generate control signals grounded in visual observations and language instructions. Current VLA policies can be broadly categorized into two directions: (1) diffusion/flow-matching policies\cite{pi_0, pi_05, smolvla, nora, evo-1}, which produce smooth actions but require slower inference and large-scale pretraining, and (2) auto-regressive policies \cite{openvla-oft, spatialforcing, vla-adapter}, which exhibit stronger in-domain fitting capability but often struggle with tasks requiring fine-grained perception and robust manipulation. In this paper, we seek to push the limit of the second path.

In the context of auto-regressive policies, we argue that one core bottleneck is not the choice of encoder -- i.e., the representation capacity of the selected backbone -- but rather how the extracted visual-language information is utilized, that is, the efficiency with which the policy leverages the encoder’s features. 
As shown in Fig.~\ref{fig:model_comparison}, previous auto-regressive VLAs \cite{openvla-oft, vla-adapter} often neglect the utilization of visual tokens. OpenVLA-OFT adopts parallel decoding for speedup but omits visual features during action generation, decoding action queries directly via a simple MLP. VLA-Adapter introduces an efficient design to incorporate VLM information into the policy, yet its mixed attention mechanism inadvertently creates a "structural shortcut". This bias encourages the model to derive task-relevant signals primarily from learnable action queries while largely bypassing the critical spatial details embedded in visual features. This neglect of visual details allows the policy to perform basic tasks via coarse action queries, but limits its performance in scenarios that demand precise and sensitive manipulation.

To investigate the roots of the problem in auto-regressive policy, we perform a series of controlled experiments to diagnose why current VLA models fail to bridge the gap from vision to action effectively. Our analysis reveals three fundamental challenges:
\begin{itemize}
    \item \textbf{Architectural Bias:} Structural shortcuts in mixed attention enable the model to bypass concrete visual details by favoring the easier action query pathway, causing the model to overlook fine-grained visual details and produce less precise actions.
    \item \textbf{Information Overload:} The excessive quantity of visual tokens dilutes the model's attention, making it difficult for the policy to concentrate on critical manipulation regions.
    \item \textbf{Task-Irrelevant Noise:} Abundant background information results in a low signal-to-noise ratio (low quality) during cross-modal interactions, where meaningful task-relevant signals are buried under environmental noise, hindering action accuracy.
\end{itemize}

Motivated by these insights, we propose \textbf{FocusVLA}, a new VLA paradigm designed to explicitly drive the attention on task-relevant visual details. FocusVLA addresses the aforementioned bottlenecks through two core innovations:
\begin{enumerate}
\item \textbf{Modality Cascaded Attention} to eliminate structural shortcuts. We replace the mixed attention in VLA-Adapter with a cascaded attention mechanism to sever the shortcut pathways within the policy. By enabling action to retrieve visual information independently, it forces the model to extract task-critical details from the visual features. As a result, Fig.~\ref{fig:vis_attn_map} illustrates a distinct transition in the attention distribution, shifting from a distracted, image-wide pattern to a highly focused and task-specific one.
\item \textbf{Focus Attention} to address the quantity and quality limitations. To manage the excessive quantity, we introduce Patch-level Focus that prunes redundant and task-irrelevant visual patches, effectively improving manipulation accuracy by reducing distracting information. To enhance signal quality, we introduce Channel-level Focus, an attention gating module to explicitly suppress task-irrelevant background noise, thereby strengthening the model's instruction-following capability.
\end{enumerate}
Together, these components significantly enhance the model's ability to deeply exploit task-relevant information within visual tokens, leading to more precise and robust manipulation.

The effectiveness of FocusVLA is validated through extensive evaluations on both simulated benchmarks and real-world robotic tasks. Specifically, our approach achieves state-of-the-art performance across the LIBERO \cite{libero} and RoboTwin \cite{robotwin} benchmarks. In challenging fine-grained manipulation tasks, such as the "Hugging Mug" task in RoboTwin, existing models often struggle to maintain consistent success, whereas FocusVLA demonstrates a clear and significant performance advantage. Furthermore, our approach significantly accelerates training convergence, achieving a $1.5\times$ overall speedup compared to VLA-Adapter on LIBERO, and a remarkable $5\times$ speedup specifically on LIBERO-Spatial. These results underscore that focusing on task-relevant visual details leads to superior manipulation precision and training efficiency.

Our contributions are summarized as follows:
\begin{itemize}
    \item We identify and validate three foundational bottlenecks in visual utilization of auto-regressive policies, and reveal that performance is primarily constrained by inefficient visual information usage rather than the intrinsic quality of representations.
    \item We propose Modality Cascaded Attention to eliminate structural bias in policy networks, enabling actions to independently retrieve task-relevant visual information and transforming attention patterns from distracted to focused.
    \item We propose Focus Attention, a dual-level mechanism that prunes redundant visual patches and suppresses background noise, thereby providing richer and more task-relevant visual details for precise manipulation.
\end{itemize}
\section{Related Work}

\subsection{General Vision-Language-Action Paradigms}
VLA models can be broadly categorized into two policy paradigms: auto-regressive policies and diffusion/flow-matching-based policies. Auto-regressive policies exhibit strong in-domain performance but often struggle with precise manipulation. Recent approaches have shifted from decoding discrete action tokens via detokenizers \cite{rt1, rt2, openvla, octo} to predicting continuous action chunks using MLP-based \cite{act, robocat, openvla-oft, rth, starvla} policies. Building on this trend, \cite{vla-adapter} further scales the paradigm by introducing a transformer-based policy that efficiently bridges information from the VLM to the control policy. Another line of work employs diffusion/flow-matching-based policies \cite{diffusionpolicy, flowvla, xvla, pi_0, pi_05, internvla-m1, st4vla}, using stochastic denoising to model complex action distributions; however, their high sampling cost limits real-time responsiveness. Our research focuses on auto-regressive policies and builds upon \cite{vla-adapter}.

\subsection{Enhanced Representations in VLA}
Enhancing visual representations and utilization are two key directions in VLA research. For representations, prior works \cite{spatialvla, 3dvla, qdepthvla, yuan2025depthvla, pointvla, falcon} leverage explicit depth, point clouds, or 3D features from \cite{vggt} to provide additional spatial information. For utilization, approaches such as \cite{lightvla, vlapruner, specprune, vlacache} remove redundant tokens to improve inference efficiency, while \cite{reconvla, starvla} incorporate additional information to provide stronger supervision, significantly boosting performance while introducing no inference cost. In this work, we systematically investigate the impact of enhanced visual representations and improved visual utilization in the same experimental setup.

\begin{figure}[t]
  \centering
  \includegraphics[width=1.\linewidth]{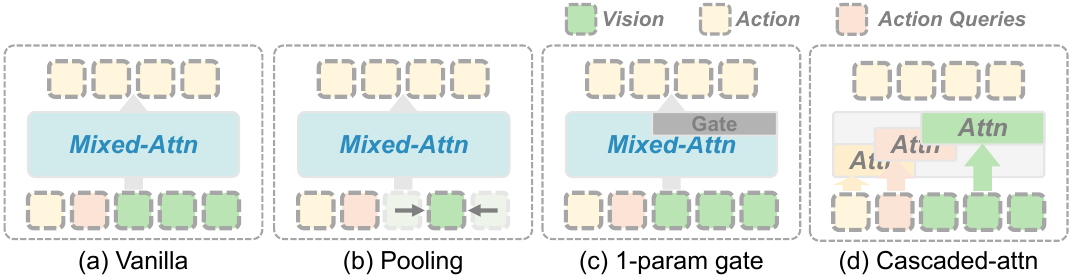}
  \vspace{-5mm}
  \caption{Architectures of the four proposed policy variants. (a) Vanilla: A baseline utilizing visual tokens without constraints. (b) Pooling: Patch-level optimization by reducing visual token count via 2×2 pooling. (c) 1-param gate: Channel-level optimization that attenuates visual signal intensity using a single-parameter gate. (d) Cascaded attention: Structure-level optimization that alters feature interaction patterns through cascaded multi-head attention.}
  \vspace{-3mm}
  \label{fig:model_analysis}
\end{figure}

\section{Methodology}
\subsection{Preliminary}
VLA-Adapter \cite{vla-adapter} integrates two VLM conditions, vision features $C^V_t$, and action queries $C^{AQ}_t$, into the action latent $A_t$, where $t$ denotes that the conditions are taken from the $t$-th VLM layer and injected into the $t$-th policy layer. Each attention layer computes a mixed attention output. First, the attention weights $W_t$ are calculated as follows:
\begin{equation}
{W}_t = \text{Softmax} \left ( \frac{\left[ {S}_V \odot \text{Tanh(g)}, \; {S}_{AQ}, \; {S}_{A} \right]}{\sqrt{d}} \right),
\end{equation}
where $[.]$ denotes the concatenation operation, $\odot$ denotes dot product, $g$ denotes its proposed single-parameter gate, and $S$ represent the attention scores:
\begin{align}
{S}_V & = (\sigma_q(A_t)) (\sigma_k(C^V_t))^\top, \\
{S}_{AQ} & = (\sigma_q(A_t)) (\sigma_k(C^{AQ}_t))^\top, \\
{S}_{A} & = (\sigma_q(A_t)) (\sigma_k(A_t))^\top.
\end{align}
Here, $\sigma$ denotes MLP projection. The aggregated action latent is then computed:
\begin{equation}
\hat{A_t} = W_t (\sigma_v([C^V_t, C^{AQ}_t, A_t]))^\top.
\end{equation}
Finally, $\hat{A_t}$ is passed through a residual Feed-Forward Network (FFN) to produce the updated latent $A_{t+1}$. Repeating this for N layers yields the final action chunk.

\begin{figure*}[t] 
  \centering
  
  \begin{subfigure}{0.24\textwidth}
    \centering
    \includegraphics[width=\linewidth]{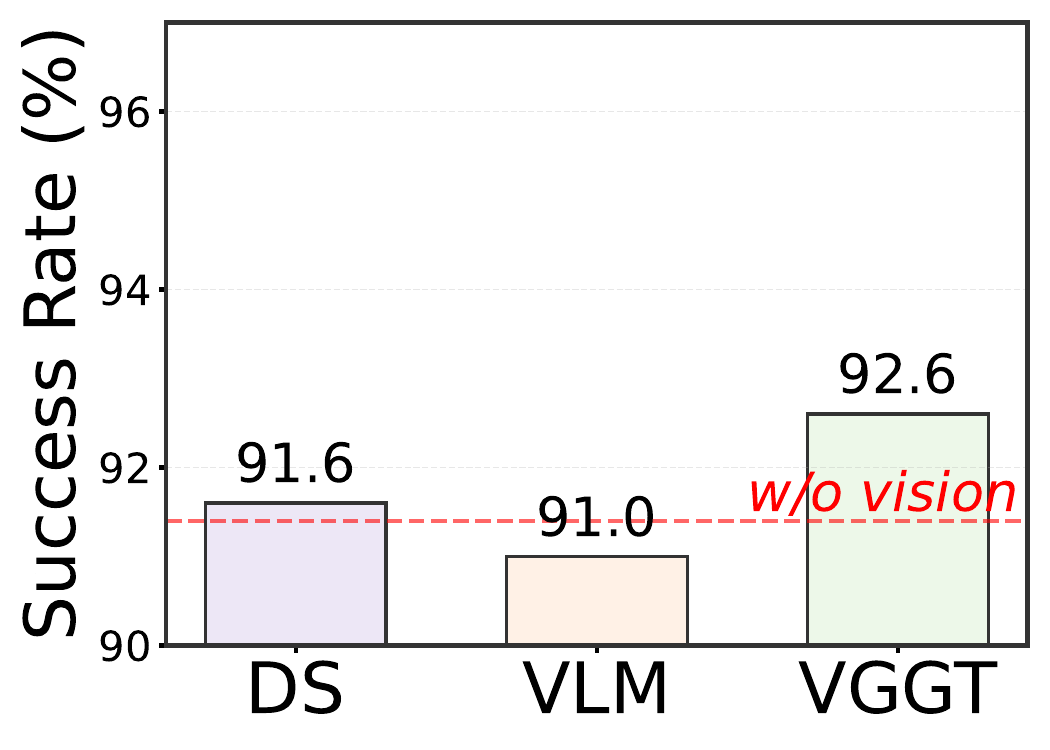}
    \caption{Vanilla}
    \label{fig:res_vanilla}
  \end{subfigure}
  \hfill
  \begin{subfigure}{0.24\textwidth}
    \centering
    \includegraphics[width=\linewidth]{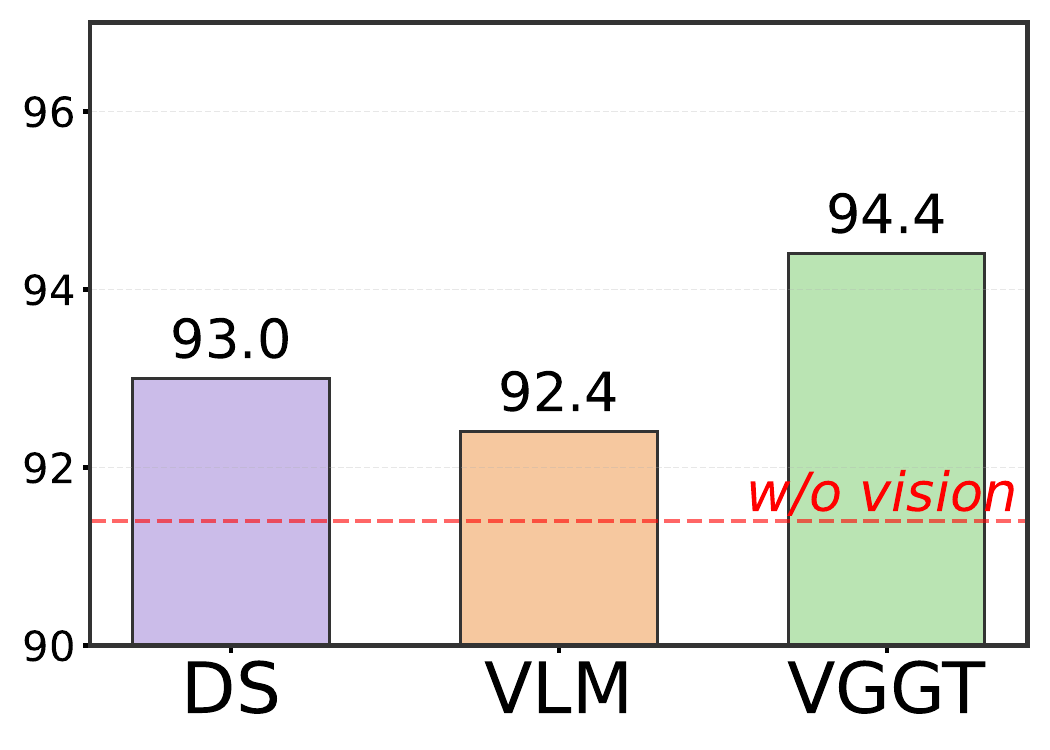}
    \caption{Pooling}
    \label{fig:res_pooling}
  \end{subfigure}
  \hfill
  \begin{subfigure}{0.24\textwidth}
    \centering
    \includegraphics[width=\linewidth]{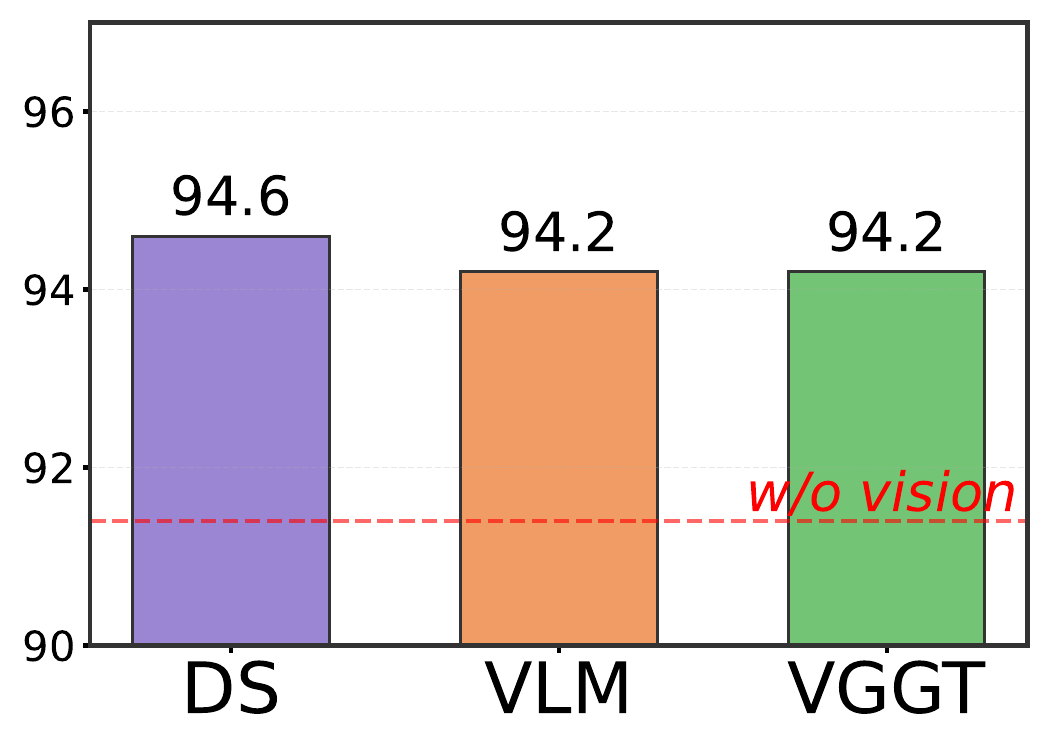}
    \caption{1-param gate}
    \label{fig:res_gate}
  \end{subfigure}
  \hfill
  \begin{subfigure}{0.24\textwidth}
    \centering
    \includegraphics[width=\linewidth]{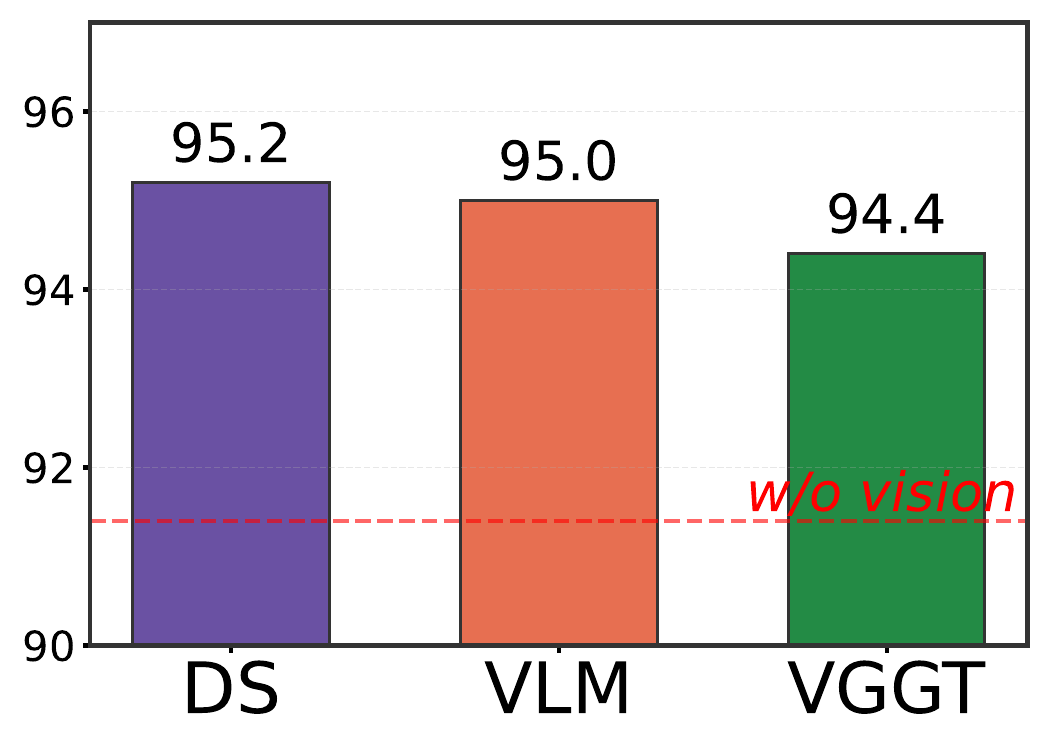}
    \caption{Cascaded-attn}
    \label{fig:res_cascaded}
  \end{subfigure}
  \vspace{-1mm}
  \caption{Success Rates of different policy architectures on the LIBERO-Long. We evaluate four different structures across three distinct visual representations as detailed in Section \ref{sec:analysis}. The experimental results reveal that existing policies often suffer from three critical biases: token quantity imbalance, low signal-to-noise ratio, and structural bias. By systematically addressing these issues through our proposed constraints, we achieve consistent and significant performance improvements across all visual representations. These findings demonstrate that VLA performance is primarily limited by how visual information is utilized, rather than by the quality of visual representations. “VLM” denotes the output visual features from the PrismaticVLM trained on Qwen2.5-0.5B \cite{qwen2}, while “DS” denotes the combined features of DINOv2 and SigLIP \cite{dinov2,siglip}. The red dashed line indicates the performance of a variant that does not use VLM visual features, relying solely on action queries.}
  \vspace{-3mm}
  \label{fig:res_analysis}
\end{figure*}

\begin{figure}[t]
  \centering
  \includegraphics[width=1.\linewidth]{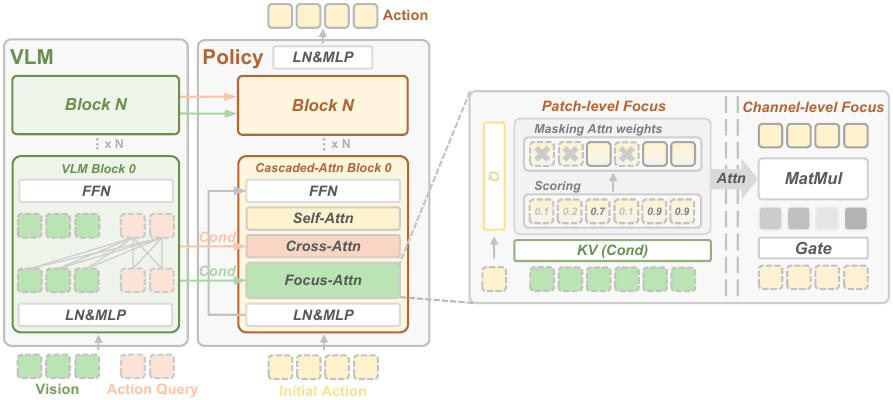}
  \vspace{-5mm}
  \caption{The architecture of FocusVLA. Our policy explicitly addresses visual utilization bottlenecks through two core components: (1) Cascaded Attention: By enabling the action latent to query each modality independently, the model is forced to focus on and extract only the most task-relevant visual details, ensuring a more targeted feature utilization; and (2) Focus Attention: To resolve the quantity and quality limitations, this component employs patch-level pruning to discard irrelevant tokens and channel-level gated suppression to mitigate feature noise.
  }
  \vspace{-3mm}
  \label{fig:method}
\end{figure}

\subsection{How to Utilize Visual Representation?}
\label{sec:analysis}
Although numerous vision-language-to-action (VL-to-A) paradigms have been proposed in prior works \cite{act, robocat, openvla-oft, rth, starvla}, the depth of visual utilization within these frameworks has not been extensively explored. In this section, we investigate the fundamental bottlenecks that hinder VLA models from efficiently utilizing visual information. In summary, we mainly focus on the following research questions:
\begin{itemize}
    \item \textbf{Question 1.1:} Do raw visual representations provide a positive contribution to action generation, and how can this potential be fully realized?
    \item \textbf{Question 1.2:} Is the mechanism for utilizing visual representations more critical than the intrinsic quality of the representations themselves?
\end{itemize}

\noindent \textbf{Experimental Setting. }
We evaluate four frameworks with three visual representations. For \textbf{Question 1.1}, we introduce a baseline that utilizes visual tokens without any constraints, as shown in Fig.~\ref{fig:model_analysis} (a). Subsequently, we compare three types of constraints: reducing the number of visual tokens via simple 2×2 pooling (patch-level optimization), attenuating visual signal intensity with a single-parameter gate (channel-level optimization), and modifying attention patterns through cascaded attention (structure-level optimization), as illustrated in Fig.~\ref{fig:model_analysis} (b), (c), and (d), respectively.
For \textbf{Question 1.2}, we compare three distinct types of visual representations across the above structures—DINOv2+SigLIP \cite{dinov2, siglip} (2D features without task-relevant information), VLM \cite{qwen2} output (2D features containing task-relevant information), and VGGT \cite{vggt} (implicit 3D information without task-relevant information).
Their performances are evaluated on LIBERO-Long, the most complex benchmark of LIBERO, and reported in Fig.~\ref{fig:res_analysis}. 
From the results, we draw the following key findings:


\textbf{Key Finding 1. }
Simply reducing the number of visual tokens or suppressing visual signal intensity with a single-parameter gate (converging to near-zero) can significantly improve performance, indicating that VLA policies suffer from both quantity imbalance and low signal-to-noise ratio: An excessive number of visual tokens overwhelm attention distributions, while most of them contribute little task-relevant information.


\textbf{Key Finding 2. }
Transitioning from mixed to cascaded attention eliminates structural bias and shifts the attention distribution from distracted to focused (as shown in Fig.~\ref{fig:vis_attn_map}), producing the largest performance gain and indicating that focus is the key driver of effective visual utilization.



\textbf{Key Finding 3. }
Different visual representations perform poorly when used naively, but once visual utilization is properly regulated, all representations achieve significant performance improvements. This indicates that VLA performance is primarily constrained by how visual information is utilized, rather than by the inherent quality of the visual representations.

\subsection{FocusVLA Methodology}

\textbf{Overall.} Motivated by our analysis, we propose a policy architecture that explicitly addresses the bottlenecks mentioned above.
Our method incorporates Modality Cascaded Attention to mitigate structural bias, and Focus Attention to regulate visual information at both patch and channel levels, addressing the quantity and quality limitations. A detailed overview of the proposed FocusVLA framework is presented in Fig.~\ref{fig:method}.

\noindent \textbf{Modality Cascaded Attention.}
To eliminate the structural bias, instead of mixing all modalities within a shared attention distribution, we decouple their interactions through a cascaded design. At each layer, the action latent $A_t$ ($t$ denotes the $t$-th layer) sequentially integrates single-modality information:
\begin{align}
 H_A = \text{Attn}(A_t, A_t),
 H_{AQ} = \text{Attn}(A_t, C^{AQ}_t),
 H_V = \text{Attn}(A_t, C^V_t),
\end{align}

The outputs are concatenated by a fusion MLP:
\begin{equation}
\hat{A_t} = \sigma_\text{fusion}([H_A, \; H_{AQ}, \; H_V]),
\end{equation}
and then passed through a residual FFN:
\begin{equation}
A_{t+1} = \text{FFN}(\hat{A_t}) + A_t.
\end{equation}

By integrating each modality sequentially rather than mixing them, this design prevents the action latent from over-relying on any single modality, thereby mitigating structural biases and forcing the model to attend to task-relevant regions and details.

\noindent \textbf{Focus Attention.}
In particular, within $H_V = \text{Attn}(A_t, C^V_t)$, we further apply Focus Attention to address both token quantity imbalance and low signal-to-noise ratio. It operates on two complementary parts: patch-level regulation, which prunes irrelevant visual tokens, and channel-level regulation, which suppresses noisy feature channels. Together, these mechanisms ensure that the action latent receives the most informative and task-relevant visual features.

\noindent \textbf{Patch-level Focus. } 
Unlike existing methods that introduce token selection within the VLM part to accelerate inference, our Patch-level Focus is applied at the policy part and is explicitly designed to improve action performance rather than computational efficiency. This distinction is crucial: due to its large scale and extensive pretraining, the VLM can naturally align image regions with semantic concepts, whereas the VLA policy, trained from scratch with significantly fewer parameters, often fails to consistently attend to task-relevant regions, resulting in distracted and noisy visual conditioning. 

Specifically, we select task-relevant visual tokens based on cross-attention scores between action queries and vision keys:
\begin{equation}
{W}_V = \text{Softmax}\Big(\text{TopK}\big((\sigma_q(A_t)) (\sigma_k(C^V_t))^\top \big)\Big),
\end{equation}
where only the tokens with top-$K$ scores are retained for information propagation, while the other tokens are masked out. The vision-to-action attention is then computed as:
\begin{equation}
H_V = ({W}_V)\big(\sigma_v(C^V_0)\big)^\top,
\end{equation}
where we use the raw feature $C^V_0$ from the visual backbone as the vision values. While deeper VLM features are biased toward semantic information, shallow features alone may struggle to focus on task-relevant regions. Combining $C^V_t$ as keys and $C^V_0$ as values allows the model to extract fine-grained spatial details from task-relevant regions.


\begin{table*}[t]
\centering
\small
\setlength{\tabcolsep}{5pt}
\caption{Comparisons with state-of-the-art methods on LIBERO benchmark.}
\vspace{-2mm}
\begin{tabular}{l|c|ccccc}
\toprule
Method & Param & Spatial & Object & Goal & Long & Avg. \\
\midrule
\rowcolor[HTML]{EFEFEF}
\multicolumn{7}{c}{Multi-weights (one policy per task)} \\ 
\midrule
OpenVLA \cite{openvla} & 7B & 84.7 & 88.4 & 79.2 & 53.7 & 76.5 \\
OpenVLA-OFT \cite{openvla-oft} & 7B & 97.6 & 98.4 & 97.9 & 94.5 & 97.1 \\
UniVLA \cite{univla} & 7B & 96.5 & 96.8 & 95.6 & 92.0 & 95.2 \\
Spatial Forcing \cite{spatialforcing}  & 7B & 99.4 & 99.6 & \textbf{98.8} & 96.0 & 98.5 \\
SpatialVLA \cite{spatialvla} & 4B & 88.2 & 89.9 & 78.6 & 55.5 & 78.1 \\
X-VLA \cite{xvla} & 0.9B & 98.2 & 98.6 & 97.8 & \textbf{97.6} & 98.1 \\
VLA-Adapter-Pro \cite{vla-adapter} & 0.5B & \textbf{99.6} & 99.6 & 98.2 & 96.4 & 98.5 \\
FocusVLA (ours) & 0.5B & \textbf{99.6} & \textbf{100} & \textbf{98.8} & 96.2 & \textbf{98.7} \\
\midrule
\rowcolor[HTML]{EFEFEF}
\multicolumn{7}{c}{Single-weight (one shared policy for four tasks)} \\
\midrule
Pi0.5 \cite{pi_05} & 3B & 98.8 & 98.2 & \textbf{98.0} & \textbf{92.4} & 96.9 \\
NORA-1.5 \cite{nora} & 3B & 98.0 & 96.0 & 95.4 & 90.5 & 95.0 \\
EVO-1 \cite{evo-1} & 0.5B & 92.7 & 97.7 & 96.3 & 92.3 & 94.8 \\
VLA-Adapter-Pro \cite{vla-adapter} & 0.5B & 98.8 & 96.2 & 95.6 & 91.6 & 95.6 \\
FocusVLA (ours) & 0.5B & \textbf{99.2} & \textbf{98.4} & 97.0 & \textbf{92.4} & \textbf{97.0} \\
\bottomrule
\end{tabular}
\vspace{-3mm}
\label{tab:libero_results}
\end{table*}

\noindent \textbf{Channel-level Focus. }
Since our method realizes a focused attention pattern to concentrate on task-relevant regions, uniformly suppressing all visual information via a single parameter gate is less effective.
To address this, we replace VLA-Adapter's \cite{vla-adapter} single-parameter gate with the element-wise gate from Gated Attention \cite{gatedattention}, enabling finer-grained noise control that suppresses task-irrelevant channels while preserving task-relevant visual signals.

Specifically, following \cite{gatedattention}, we enhance the quality of vision-to-action attention by applying an adaptive gate to attention outputs $H_V$:
\begin{equation}
H_V' = H_V \odot \sigma_g(A_t),
\end{equation}
where $\sigma_g$ denotes gate MLP, and $\odot$ denotes dot product. The gate adaptively suppresses noisy channels, and as observed in our experiments, also enhances the model's instruction-following ability (see Sec.~\ref{sec:sim_expe} and Fig.~\ref{fig:vis_gate}).

Together, the proposed methods transform distracted visual utilization into focused, task-aligned interactions, enabling precise and robust action generation.


\section{Simulation Experiments}
\label{sec:sim_expe}
\subsection{Experimental Setup}

\noindent \textbf{Benchmark.}
We evaluate our method on two widely used simulation benchmarks: LIBERO~\cite{libero} and RoboTwin \cite{robotwin}. LIBERO comprises four task suites -- Spatial, Object, Goal, and Long -- each providing 500 expert demonstrations across 10 tasks to assess policy generalization over diverse spatial layouts, objects, goals, and long-horizon scenarios. RoboTwin is a real-to-sim bimanual benchmark with an easy in-domain setting and a more challenging domain-randomized setting. We evaluate on a diverse set of tasks and report success rate (SR) as the evaluation metric for both benchmarks.

\begin{figure}[t]
  \centering
  \includegraphics[width=1.\linewidth]{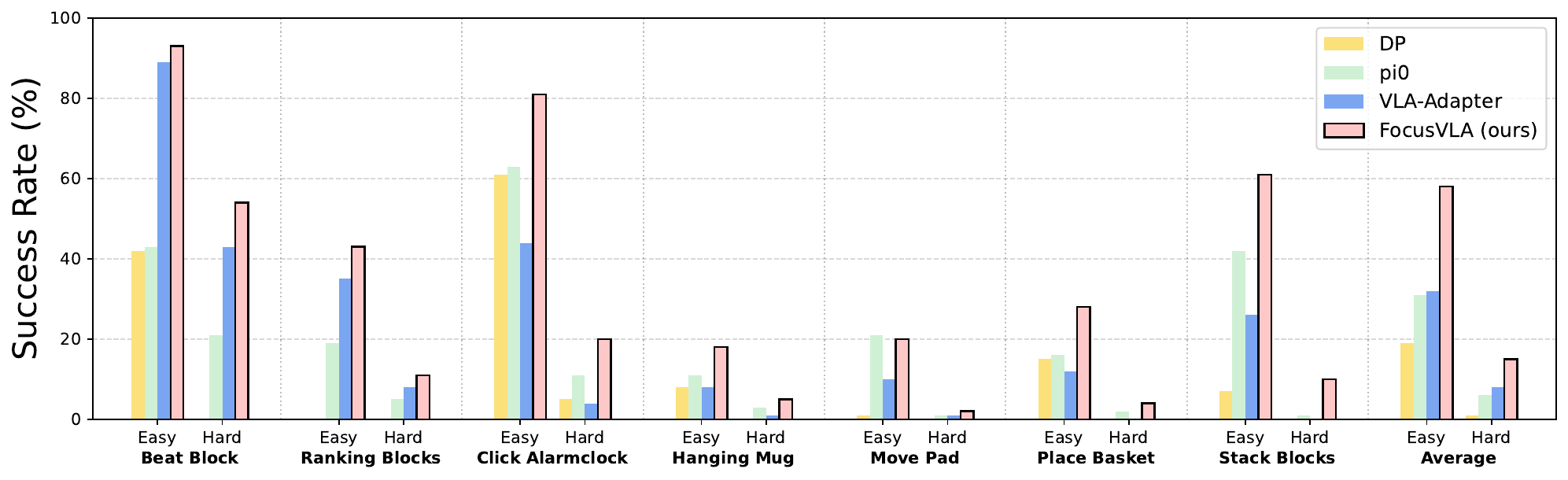}
  \vspace{-4mm}
  \caption{Comparisons with state-of-the-art methods on RoboTwin 2.0 benchmark.}
  \vspace{-4mm}
  \label{fig:res_robotwin}
\end{figure}

\noindent \textbf{Implementation Details.}
Following \cite{vla-adapter}, we adopt DINOv2 and SigLIP \cite{dinov2, siglip} as the visual backbones, and use PrismaticVLM trained on Qwen2.5-0.5B \cite{qwen2} as the VLM backbone. Unless otherwise specified, all hyperparameters follow the settings in \cite{vla-adapter}.

For LIBERO, we train FocusVLA using 4 NVIDIA A100 GPUs with a total batch size of 64. 
Under the single-weight setting (a shared policy across four tasks), the model is trained for 100k steps, with evaluation conducted every 10k steps. 
Under the multi-weights setting (one policy per task), we train for 25k steps on LIBERO-Spatial, 10k steps on LIBERO-Object, 50k steps on LIBERO-Goal, and 80k steps on LIBERO-Long, performing evaluation every 5k steps. Each suite is evaluated for 500 trials.

For RoboTwin, we use 8 A100 GPUs with a total batch size of 64. 
The model is trained for 20k steps under the Easy setting and 100k steps under the Hard setting, with evaluation carried out every 10k steps. Each easy task is evaluated for 100 trials, and each hard task is evaluated for 300 trials under random seeds.

\subsection{Comparison with State-of-the-art Methods}

\noindent \textbf{LIBERO.} We evaluate FocusVLA on the LIBERO benchmark across four task suites. Tab.~\ref{tab:libero_results} demonstrates that FocusVLA achieves state-of-the-art performance under both single-weight and multi-weight settings.

Specifically, under the multi-weight setting, FocusVLA achieves an average success rate of 98.7\%. Remarkably, despite having only 0.5B parameters, it surpasses substantially larger 7B-scale models such as OpenVLA-OFT \cite{openvla-oft} and Spatial Forcing \cite{spatialforcing}. This result highlights a small-but-powerful property: with significantly fewer parameters, our model exhibits strong task-specific mastery, and precise and fine-grained manipulation capability. 

Furthermore, under the more challenging single-weight setting, which evaluates unified multi-task capacity, FocusVLA continues to show outstanding performance without any robotic pretraining. It achieves an average success rate of 97.0\%, significantly outperforming other 0.5B models such as EVO-1 \cite{evo-1} (94.8\%) and VLA-Adapter \cite{vla-adapter} (95.6\%). 

We attribute these gains to more efficient visual utilization. By leveraging visual representations in a more structured and focused manner, FocusVLA achieves higher action precision and stronger behavioral fitting capacity, which ultimately translates into consistently improved success rates across both task-specific and unified multi-task settings.

\noindent \textbf{RoboTwin. } RoboTwin evaluates a more fine-grained and comprehensive set of robot manipulation tasks. Fig.~\ref{fig:res_robotwin} shows that FocusVLA significantly outperforms baseline models, including DP \cite{diffusionpolicy}, pi0 \cite{pi_0}, and VLA-Adapter \cite{vla-adapter}, across diverse tasks. In tasks that require precise manipulation, such as Hugging Mug, FocusVLA excels by leveraging rich visual details to achieve higher manipulation precision, whereas VLA-Adapter struggles due to its distracted attention distribution. With this advantage, even in tasks that emphasize reasoning, such as Ranking Block, FocusVLA also achieves higher success rates.

\subsection{Ablation Study}

\definecolor{colorAttn}{HTML}{FFEDED} 
\definecolor{colorToken}{HTML}{EFFFF0} 
\definecolor{colorGate}{HTML}{FFFFE0}  
\definecolor{colorRep}{HTML}{F0F5FF}   
\definecolor{colorAttn}{HTML}{FFD6D6} 
\definecolor{colorToken}{HTML}{D7FAD9} 
\definecolor{colorGate}{HTML}{FFFACD}  
\definecolor{colorRep}{HTML}{E0E9FF}   

\begin{table}[t]
\centering
\caption{Component Analysis on LIBERO benchmark under the multi-weights setting. *DS denotes the combination of DINOv2 and SigLIP.}
\vspace{-2mm}
\label{tab:ablation_study}
\small
\setlength{\tabcolsep}{4pt}
\small
\resizebox{\columnwidth}{!}{
\begin{tabular}{lccc|ccccc}
\toprule
\multicolumn{1}{c}{\begin{tabular}[c]{@{}c@{}}\cellcolor{colorAttn}Attention\\ \cellcolor{colorAttn}Pattern\end{tabular}} & 
\begin{tabular}[c]{@{}c@{}}\cellcolor{colorToken}Patch-level\\ \cellcolor{colorToken}Focus\end{tabular} & 
\begin{tabular}[c]{@{}c@{}}\cellcolor{colorGate}Channel-level\\ \cellcolor{colorGate}Focus\end{tabular} & 
\begin{tabular}[c]{@{}c@{}}\cellcolor{colorRep}Visual\\ \cellcolor{colorRep}Representation\end{tabular} &
\begin{tabular}[c]{@{}c@{}}Spatial\\ SR (\%)\end{tabular} & 
\begin{tabular}[c]{@{}c@{}}Object\\ SR (\%)\end{tabular} & 
\begin{tabular}[c]{@{}c@{}}Goal\\ SR (\%)\end{tabular} & 
\begin{tabular}[c]{@{}c@{}}Long\\ SR (\%)\end{tabular} & 
\begin{tabular}[c]{@{}c@{}}AVG.\\ (\%)\end{tabular} \\ \midrule

\cellcolor{colorAttn} Mixed & 512 & w/o gate & VLM & 94.4 & 95.6 & 93.2 & 91.0 & 93.6 \\
\cellcolor{colorAttn} Cascaded & 512 & w/o gate & VLM & \textbf{98.0} & \textbf{98.6} & \textbf{96.2} & \textbf{95.0} & \textbf{97.0} \\ \midrule

Cascaded & \cellcolor{colorToken} 128 & w/o gate & VLM & 96.8 & 99.0 & 95.4 & 93.8 & 96.3 \\
Cascaded & \cellcolor{colorToken} 256 & w/o gate & VLM & \textbf{99.0} & 99.2 & \textbf{98.2} & \textbf{95.4} & \textbf{98.0} \\
Cascaded & \cellcolor{colorToken} 384 & w/o gate & VLM & 98.4 & \textbf{99.6} & 97.8 & 95.2 & 97.8 \\
Cascaded & \cellcolor{colorToken} 512 & w/o gate & VLM & 98.0 & 98.6 & 96.2 & 95.0 & 97.0 \\ \midrule

Cascaded & 256 & \cellcolor{colorGate} w/o gate & VLM & 99.0 & 99.2 & 98.2 & 95.4 & 98.0 \\
Cascaded & 256 & \cellcolor{colorGate} 1-param & VLM & 98.0 & 99.0 & 97.8 & 94.2 & 97.3 \\
Cascaded & 256 & \cellcolor{colorGate} Head-wise & VLM & 97.4 & \textbf{99.6} & 96.8 & 94.4 & 97.1 \\
Cascaded & 256 & \cellcolor{colorGate} Element-wise & VLM & \textbf{99.6} & 99.2 & \textbf{98.2} & \textbf{95.6} & \textbf{98.2} \\ \midrule

Cascaded & 256 & Element-wise & \cellcolor{colorRep} VLM & 99.6 & 99.2 & 98.2 & 95.6 & 98.2 \\
Cascaded & 256 & Element-wise & \cellcolor{colorRep} DS* & 99.4 & 99.4 & 98.6 & 96.0 & 98.4 \\
Cascaded & 256 & Element-wise & \cellcolor{colorRep} VGGT & 97.4 & 98.6 & 96.4 & 94.8 & 96.8 \\
Cascaded & 256 & Element-wise & \cellcolor{colorRep} VLM+DS* & \textbf{99.6} & \textbf{100.0} & \textbf{98.8} & \textbf{96.2} & \textbf{98.7} \\ \bottomrule
\end{tabular}
}
\vspace{-3mm}
\end{table}

\noindent \textbf{Component-wise Analysis. }We conduct comprehensive ablation studies on LIBERO to investigate the impact of each component in FocusVLA. As summarized in Tab.~\ref{tab:ablation_study}, we evaluate our design choices in four dimensions:

\noindent \textbf{1) Attention Pattern:} We first compare the performance of mixed attention and our proposed cascaded attention. As illustrated in the attention map visualization (Fig.~\ref{fig:vis_attn_map}), the attention distribution shifts from distracted to focused under the cascaded design, enabling the model to better capture task-relevant visual features. Consequently, the average success rate is significantly improved from 93.6\% to 97.0\%.

\noindent \textbf{2) Patch-level Focus:} We further analyze the effect of varying the number of visual tokens in our patch-level design. When using 512 tokens (i.e., without any constraint), the excessive number of visual tokens introduces substantial redundancy. The policy struggles to concentrate on task-relevant regions, leading to distracted attention and weak action generation. In contrast, reducing the token number to 128 imposes overly strong constraints. While this setting suppresses redundancy, it also removes informative visual cues that are essential for accurate manipulation, thereby degrading action quality.

\begin{wrapfigure}{r}{0.5\linewidth} 
  \centering
  \vspace{-5mm}
  \includegraphics[width=.98\linewidth]{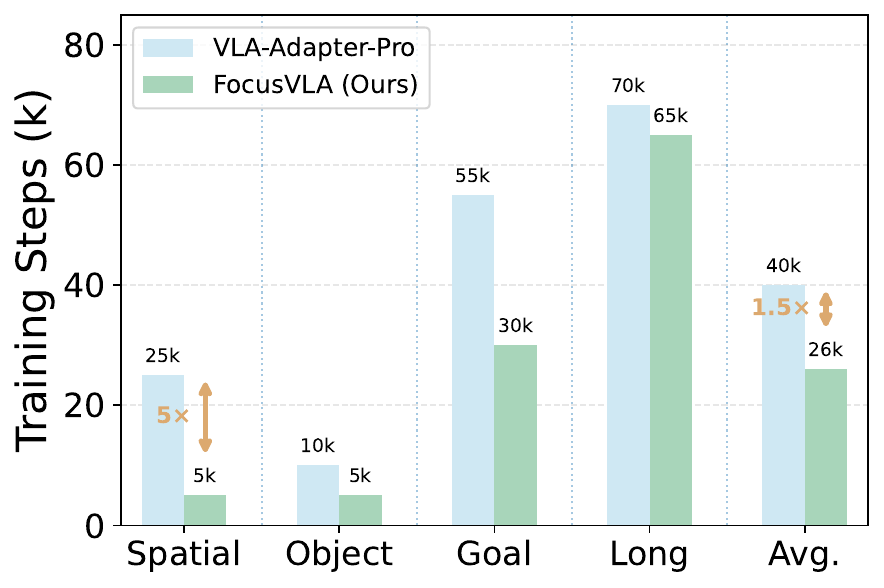}
  \vspace{-3mm}
  \caption{Comparison of training efficiency between VLA-Adapter and FocusVLA.}
  \label{fig:res_efficency}
  \vspace{-5mm}
\end{wrapfigure}

\begin{figure}[t]
    \centering
    \begin{subfigure}{0.66\textwidth}
        \centering
        \includegraphics[width=\linewidth]{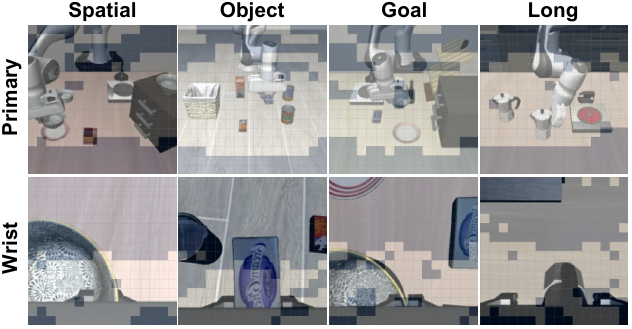}
        \caption{Effect of patch-level focus}
        \label{fig:vis_patch}
    \end{subfigure}
    \hfill 
    \begin{subfigure}{0.32\textwidth}
        \centering
        \includegraphics[width=\linewidth]{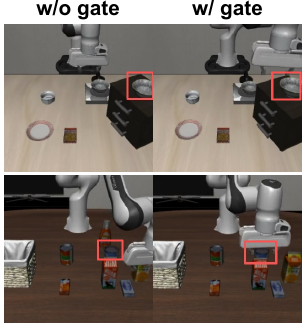}
        \caption{Effect of channel-level focus}
        \label{fig:vis_gate}
    \end{subfigure}
    \hfill
    \vspace{-3mm}
    \caption{Visualization of Focus Attention. (a) Patch-level focus. In the primary view, background regions are largely removed with key objects retained, maintaining sufficient spatial cues for policy planning. In the wrist view, representations only concentrate on the gripper and target object, enabling precise grasp execution. (b) Channel-level focus. Without gating, noisy information may lead to instruction-inconsistent actions; with gating, irrelevant signals are suppressed, yielding more accurate and instruction-consistent behavior.}
    \vspace{-5mm}
    \label{fig:vis_focus_attention}
\end{figure}

As shown in Fig.~\ref{fig:vis_patch}, our default setting with 256 visual tokens achieves a balanced focus. In the primary view, background regions are largely removed while most objects remain, preserving sufficient semantic cues for planning and decision-making. In the wrist view, the representation is further concentrated on the gripper and the target object, ensuring precise and reliable grasp execution.

\noindent \textbf{3) Channel-level Focus:} We compare different gating granularities and find that the element-wise gate performs the best, achieving a 98.2\% average success rate. This result highlights the necessity of fine-grained modulation for complex robotic manipulation, where subtle cross-modal interactions critically affect action quality.

Beyond quantitative gains, we also observe an additional benefit of gating during evaluation. As shown in Fig.~\ref{fig:vis_gate}, the model with gating can effectively filter out task-irrelevant signals. Without gating, the policy may generate actions that deviate from the instruction due to noisy or distracting information. With gating enabled, irrelevant signals are suppressed, leading to more accurate and instruction-consistent behavior.

\noindent \textbf{4) Visual Representation:} Finally, we analyze the impact of different visual backbones and examine the strengths and limitations of three types of representations. (1) VLM outputs provide rich high-level semantic information. However, prior studies \cite{vla-adapter, smolvla} have pointed out that excessive semantic abstraction may weaken the preservation of fine-grained spatial structure. (2) DINOv2 + SigLIP (DS) \cite{dinov2, siglip}, the raw input of VLM, offers a more balanced representation, maintaining spatial structure while preserving semantic alignment capability. (3) VGGT \cite{vggt} exhibits the most powerful spatial modeling ability. However, due to architectural constraints, to avoid disrupting the pretrained VLM, VGGT features can only be injected at the policy stage. As a result, their gradients are relatively weaker and less stable, which limits their performance upper bound.

The experimental results validate that the integration of VLM’s semantic awareness with DS’s spatial fidelity creates a complementary synergy. By leveraging the high-level contextual reasoning of VLM alongside the low-level details of DS, the model effectively mitigates the inherent limitations of using either modality in isolation and achieves a 98.7\% average success rate.

\noindent \textbf{Training Efficiency Analysis.} To evaluate the convergence speed of our model, we compare the training steps required by FocusVLA and VLA-Adapter to reach their optimal performance on LIBERO. As illustrated in Fig.~\ref{fig:res_efficency}, FocusVLA demonstrates superior training efficiency across all task suites with an average 1.5$\times$ speedup. Specifically, in the LIBERO-Spatial suite, FocusVLA requires only 5k steps, achieving a 5$\times$ speedup compared to the 25k steps needed by VLA-Adapter. This efficiency gain can be attributed to the fact that VLA-Adapter relies more heavily on the action queries to provide comprehensive action information, whereas FocusVLA, beyond the action queries, is able to extract task-relevant cues directly from visual representations. By offloading part of the informational burden from the action queries to visual features, FocusVLA effectively distributes the informational bandwidth, thereby facilitating faster and more stable convergence.

\begin{figure}[t]
  \centering
  \includegraphics[width=1.\linewidth]{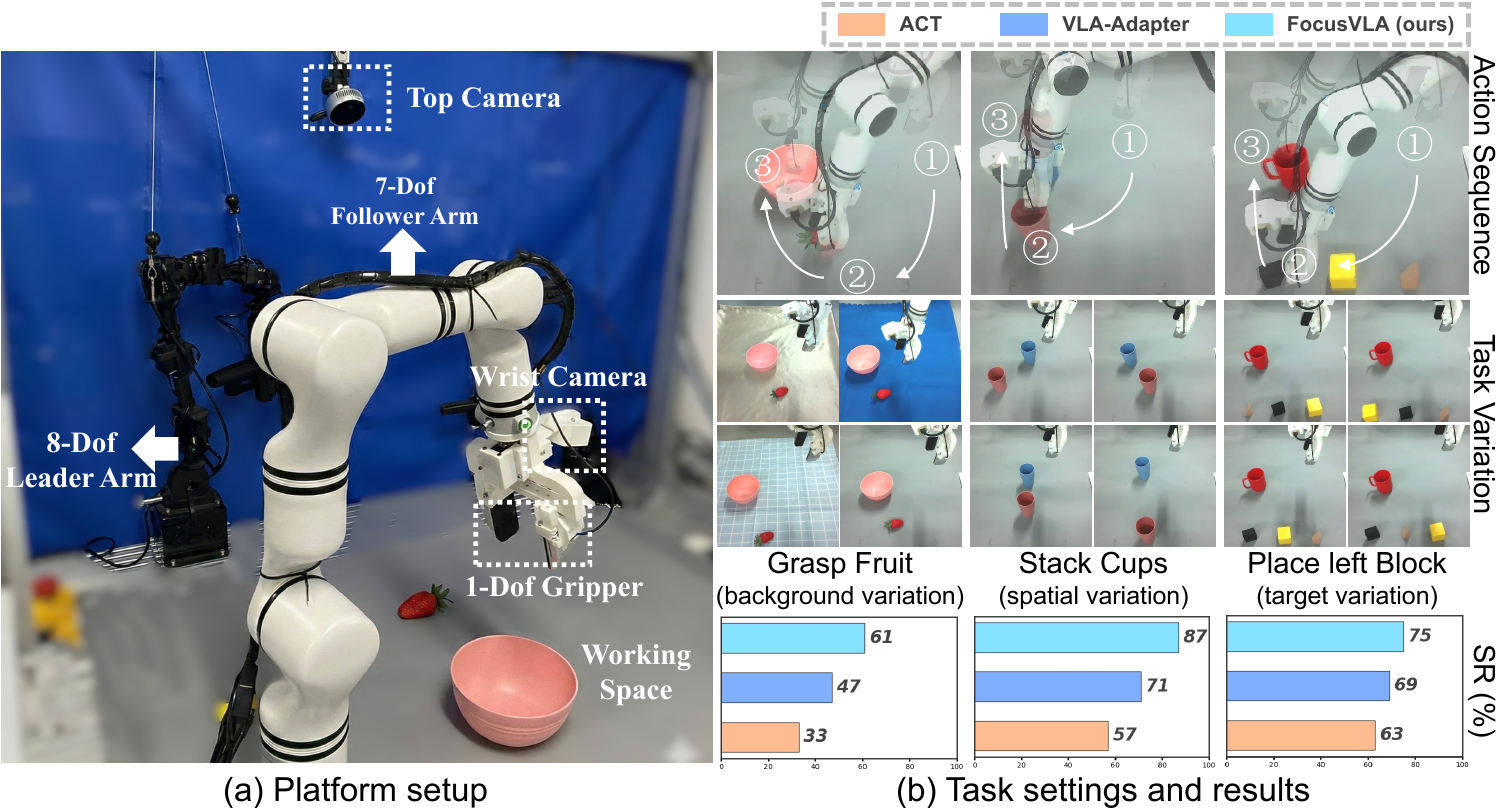}
  \vspace{-5mm}
  \caption{ Experimental settings and results in real-world scenarios.}
  \vspace{-5mm}
  \label{fig:real}
\end{figure}

\vspace{-2mm}
\section{Real-world Experiments}
\label{sec:real_expe}
\vspace{-1mm}
\subsection{Experimental Setup}
\vspace{-1mm}
As in Fig.~\ref{fig:real}, we perform real-robot experiments on the Realman platform. Each arm is equipped with a 7-DoF manipulator and a 1-DoF gripper with a top camera and a wrist camera. We design three tasks: grasping fruit under background variation, stacking cups under spatial variation, and placing the left block under target variation. All models are trained with only 50 episodes per task. For evaluation, we conduct 25 trials per variation for each task (100 trials in total).

\vspace{-2mm}
\subsection{Result Analysis}
\vspace{-1mm}

Fig.~\ref{fig:real} shows that our FocusVLA achieves higher success rates across all tasks. Specifically, in the fruit grasping task, even when the background color and texture vary, FocusVLA still reaches a high success rate, demonstrating the method’s robustness to visual changes. In the stack cups task, different spatial arrangements require the ability to focus on the correct visual regions. Our results indicate that FocusVLA effectively attends to these task-relevant regions. For the place left block task, variations in target objects demand precise manipulation due to differences in object shapes and sizes. 
Our model achieves the strongest performance due to the focus on visual details.

\vspace{-2mm}
\section{Conclusion}
\vspace{-2mm}
In this work, we revisit the vision-to-action gap in auto-regressive VLA policies and show that their performance is primarily limited by inefficient utilization of visual information rather than the representation quality of the visual model. Through systematic analysis, we identify three core bottlenecks—architectural bias, information overload, and task-irrelevant noise—that hinder precise manipulation. To address these challenges, we propose FocusVLA, which incorporates Modality Cascaded Attention to mitigate structural bias, and Focus Attention to regulate visual information at both patch and channel levels, addressing the quantity and quality limitations. By explicitly directing the model toward task-relevant visual regions while suppressing irrelevant distractions, our approach achieves more accurate, robust, and efficient action generation across both simulated and real-world benchmarks. 

\textbf{Limitations.}
While FocusVLA achieves strong performance, it still presents certain limitations.
First, although our work targets key bottlenecks in the policy side, visual utilization within the VLM side also influences action generation.
Second, scaling up either the model size or the training data is needed to further validate FocusVLA's generalization and performance.
Finally, the success rate of the real-world experiments remains constrained by the limited real-world data.
Future work in these three directions will further contribute to building a robust and generalizable robot system.

\bibliographystyle{splncs04}
\bibliography{main}

\appendix
\clearpage
\appendix

\section{More Implementation Details}

In this section, we present our implementation details and hyperparameters for FocusVLA. During training, the vision-language model (VLM) is initialized from pretrained weights and adapted using the LoRA fine-tuning scheme. In contrast, the policy module is trained from scratch with full parameter updates. All trainable parameters are optimized using the AdamW optimizer.

The backbone of FocusVLA is based on Qwen2.5-0.5B. The policy module leverages intermediate representations from the VLM, where layers 1--24 are utilized for policy learning. In addition, visual representations from DINOv2 + SigLIP and VGGT are incorporated by extracting their last-layer features. All training hyperparameters and architectural configurations are summarized in Table~\ref{tab:details}.

\begin{table}[h]
\vspace{-2mm}
\centering
\caption{Implementation details and hyperparameters of FocusVLA.}
\label{tab:details}
\begin{tabular}{lc}
\toprule
\textbf{Hyperparameter} & \textbf{Value} \\
\midrule
Backbone & Qwen2.5-0.5B \\
Number of layers $(\tau / M)$ & 24 \\
Hidden size & 896 \\
Attention heads & 8 \\

Utilization of VLM features & layers 1--24 \\
Utilization of DINOv2 + SigLIP features & last layer \\
Utilization of VGGT features & last layer \\

Visual tokens (raw) & 512 \\
Visual tokens (in Focus-Attn) & 256 \\
Number of ActionQuery tokens & 64 \\
Action chunk for single-arm & 8 \\
Action chunk for dual-arm & 16 \\
Batch size & 64 \\
Learning rate & $2\times10^{-4}$ \\
Optimizer & AdamW \\

Trainable parameters (VLM) & 103.6M \\
Trainable parameters (Policy) & 239.1M \\
Total trainable parameters & 342.7M \\

\bottomrule
\end{tabular}
\vspace{-2mm}
\end{table}

\section{More Detailed Results}

In this section, we provide detailed results of FocusVLA on both the LIBERO and RoboTwin benchmarks.

\subsection{Detailed Results on LIBERO}

Table~\ref{tab:libero_detailed} reports the task-level success rates of FocusVLA across four LIBERO task suites, including \textit{Spatial}, \textit{Object}, \textit{Goal}, and \textit{Long}. Each suite contains 10 tasks, and we report the average success rate across them.

\begin{table}[h]
\vspace{-2mm}
\centering
\caption{Task-level success rates (\%) on the LIBERO benchmark.}
\label{tab:libero_detailed}
\begin{tabular}{lccccccccccc}
\toprule
\textbf{Suite} & \textbf{1} & \textbf{2} & \textbf{3} & \textbf{4} & \textbf{5} & \textbf{6} & \textbf{7} & \textbf{8} & \textbf{9} & \textbf{10} & \textbf{Avg.$\uparrow$} \\
\midrule
Spatial & 100.0 & 100.0 & 100.0 & 100.0 & 100.0 & 100.0 & 100.0 & 96.0 & 100.0 & 100.0 & 99.6 \\
Object & 100.0 & 100.0 & 100.0 & 100.0 & 100.0 & 100.0 & 100.0 & 100.0 & 100.0 & 100.0 & 100.0 \\
Goal & 100.0 & 96.0 & 98.0 & 100.0 & 100.0 & 100.0 & 98.0 & 100.0 & 98.0 & 98.0 & 98.8 \\
Long & 96.0 & 98.0 & 100.0 & 100.0 & 98.0 & 100.0 & 94.0 & 98.0 & 84.0 & 94.0 & 96.2 \\
\bottomrule
\end{tabular}
\vspace{-2mm}
\end{table}

FocusVLA achieves near-perfect performance on the Spatial and Object suites, reaching average success rates of 99.6\% and 100.0\%, respectively. The model also performs strongly on Goal and Long tasks, demonstrating robust long-horizon reasoning and object interaction capabilities.

\subsection{Detailed Results on RoboTwin}

We further evaluate FocusVLA on the RoboTwin benchmark. Table~\ref{tab:robotwin_detailed} reports success rates under both \textit{easy} and \textit{hard} settings across multiple baseline methods.

\begin{table*}[h]
\vspace{-2mm}
\centering
\caption{Detailed success rates (\%) on RoboTwin benchmark.}
\label{tab:robotwin_detailed}
\begin{tabular}{lcccccccc}
\toprule
\multirow{2}{*}{\textbf{Task}} 
& \multicolumn{2}{c}{DP} 
& \multicolumn{2}{c}{$\pi_0$} 
& \multicolumn{2}{c}{VLA-Adapter} 
& \multicolumn{2}{c}{FocusVLA} \\
\cmidrule(lr){2-3} \cmidrule(lr){4-5} \cmidrule(lr){6-7} \cmidrule(lr){8-9}
& Easy & Hard & Easy & Hard & Easy & Hard & Easy & Hard \\
\midrule
Beat Block Hammer & 42 & 0 & 43 & 21 & 89 & 43 & 93 & 54 \\
Blocks Ranking RGB & 0 & 0 & 19 & 5 & 35 & 8 & 43 & 11 \\
Click Alarmclock & 61 & 5 & 63 & 11 & 44 & 4 & 81 & 20 \\
Hanging Mug & 8 & 0 & 11 & 3 & 8 & 1 & 18 & 5 \\
Move Pillbottle Pad & 1 & 0 & 21 & 1 & 10 & 1 & 20 & 2 \\
Place Object Basket & 15 & 0 & 16 & 2 & 12 & 0 & 28 & 4 \\
Stack Blocks Two & 7 & 0 & 42 & 1 & 26 & 0 & 61 & 10 \\
\midrule
AVG & 19 & 1 & 31 & 6 & 32 & 8 & 58 & 15 \\
\bottomrule
\end{tabular}
\vspace{-2mm}
\end{table*}

Compared with prior approaches, FocusVLA achieves the best average performance on both easy and hard settings. In particular, it significantly improves task success rates in complex manipulation tasks such as \textit{Stack Blocks Two} and \textit{Hanging Mug}. These results demonstrate the strong generalization ability of FocusVLA across diverse robotic manipulation scenarios.

\section{Qualitative Results and Case Studies}
In this section, we present qualitative results of FocusVLA on the LIBERO and RoboTwin benchmarks, as well as real-world experiments, including benchmark visualizations, comparative cases, and representative failure cases.

\subsection{Visualizations of LIBERO Tasks}
\begin{figure}[h]
\vspace{-2mm}
\centering
\includegraphics[width=\linewidth]{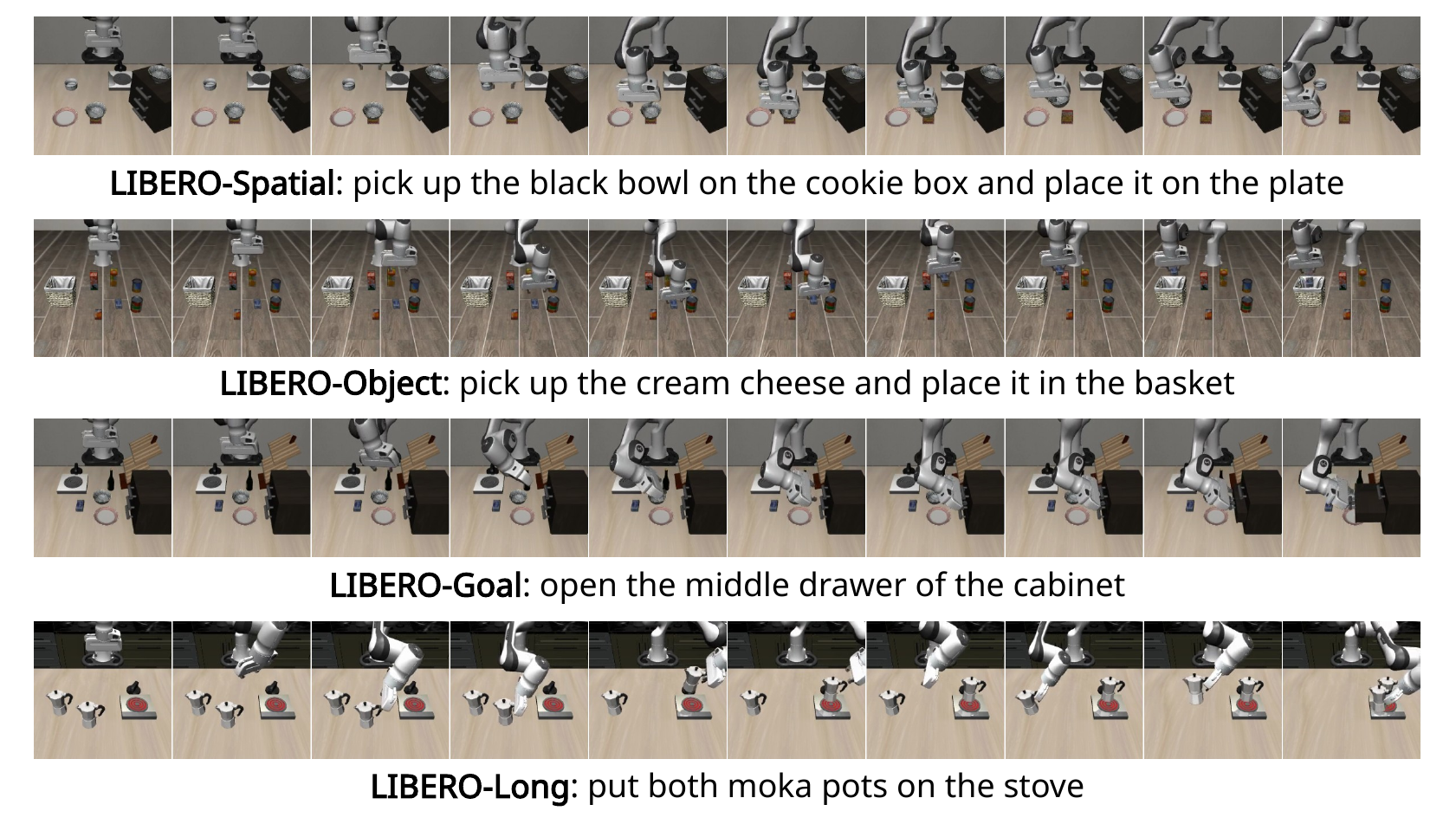}
\caption{Visualizations of FocusVLA executing tasks in the LIBERO benchmark.}
\label{fig:libero_vis}
\vspace{-2mm}
\end{figure}

Fig.~\ref{fig:libero_vis} presents qualitative examples of FocusVLA executing tasks in the LIBERO benchmark. The figure illustrates the simulated environments and provides an overview of representative tasks from the four LIBERO suites. LIBERO-Spatial evaluates the robot's ability to grasp objects according to different spatial relationships. LIBERO-Object evaluates the robot's capability to manipulate target objects with different colors and shapes. LIBERO-Goal tests the robot's ability to accurately identify and grasp the specified target object. LIBERO-Long evaluates the robot's capability to perform long-horizon tasks that require multiple sequential actions. As shown in the figure, our method achieves strong performance across all four types of tasks.

\begin{figure}[h]
\vspace{-2mm}
\centering
\includegraphics[width=\linewidth]{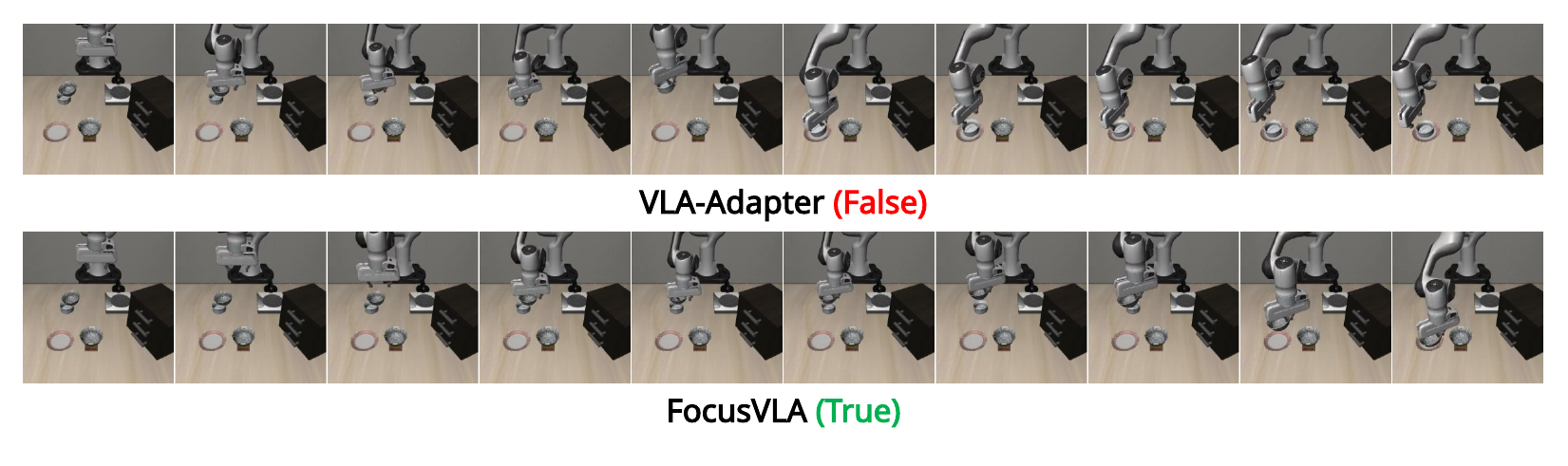}
\caption{Comparison between FocusVLA and VLA-Adapter on the same LIBERO task.}
\label{fig:libero_com}
\vspace{-2mm}
\end{figure}

Fig.~\ref{fig:libero_com} compares rollouts of our method and VLA-Adapter on the same task. Our method performs more precise manipulation, while VLA-Adapter fails to fully utilize visual features due to architectural bias. As a result, it mistakenly places an object that is visually similar to the target object at the goal location. This example highlights the importance of generating actions based on detailed visual features.

\begin{figure}[t]
\vspace{-2mm}
\centering
\includegraphics[width=\linewidth]{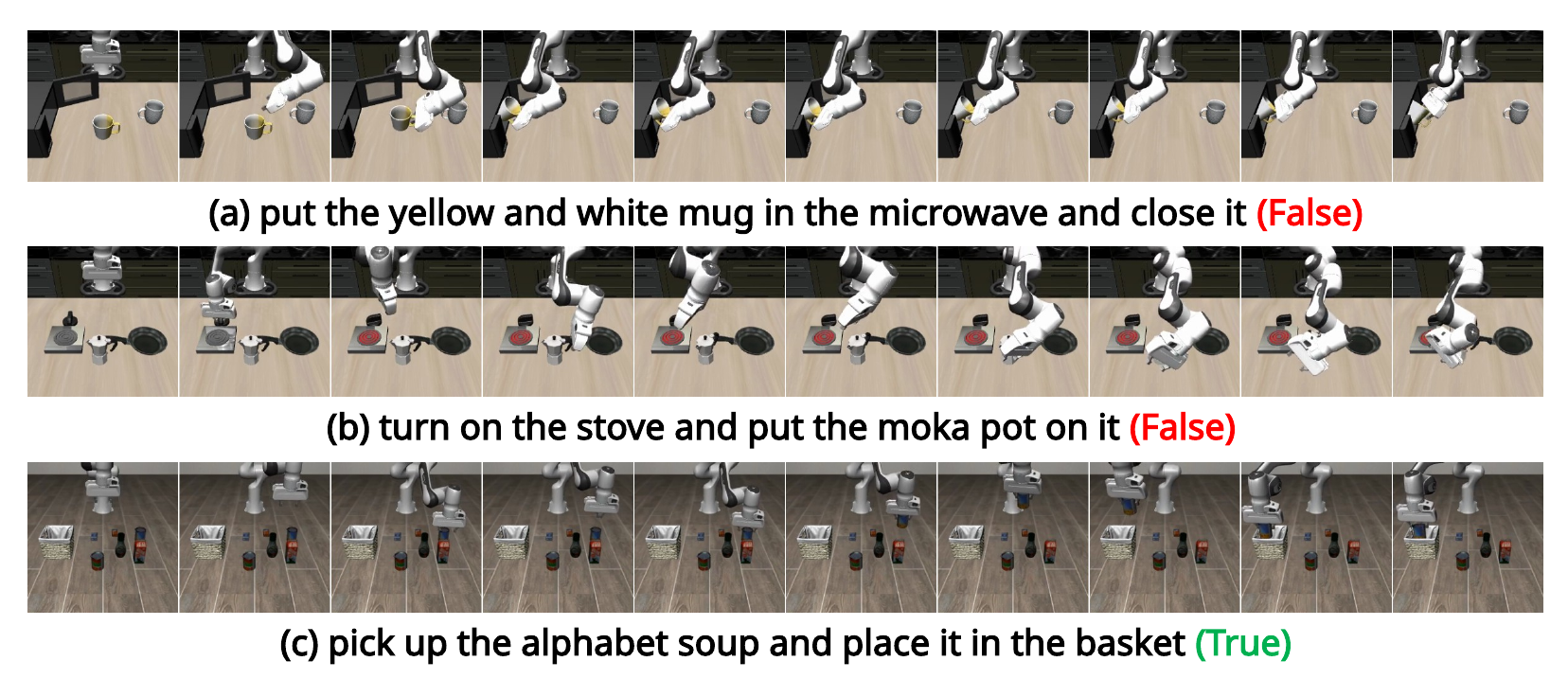}
\caption{Examples of unexpected situations during task execution in LIBERO simulation, including two failure trajectories and one successful recovery case.}
\label{fig:libero_failure}
\vspace{-2mm}
\end{figure}

Fig.~\ref{fig:libero_failure} shows several cases where unexpected situations occur during task execution, including two failure trajectories and one successful trajectory. In Fig.~\ref{fig:libero_failure}(a), an unexpected event occurs but the model does not adjust its behavior. This happens because imitation learning primarily learns optimal trajectories, and recovery from unexpected situations relies on out-of-distribution generalization. In Fig.~\ref{fig:libero_failure}(b), after the object drops, the model does not attempt to re-grasp it and instead places a nonexistent object at the target location. This behavior indicates overfitting to the optimal trajectory: while it improves manipulation precision under normal conditions, it reduces robustness and generalization when unexpected events occur. Fig.~\ref{fig:libero_failure}(c) shows a successful recovery case. Although the robot initially fails to grasp the object, the model retries the grasp and eventually completes the task successfully, demonstrating a certain level of robustness to unexpected situations.

\subsection{Visualizations of RoboTwin Tasks}

\begin{figure}[h]
\vspace{-2mm}
\centering
\includegraphics[width=\linewidth]{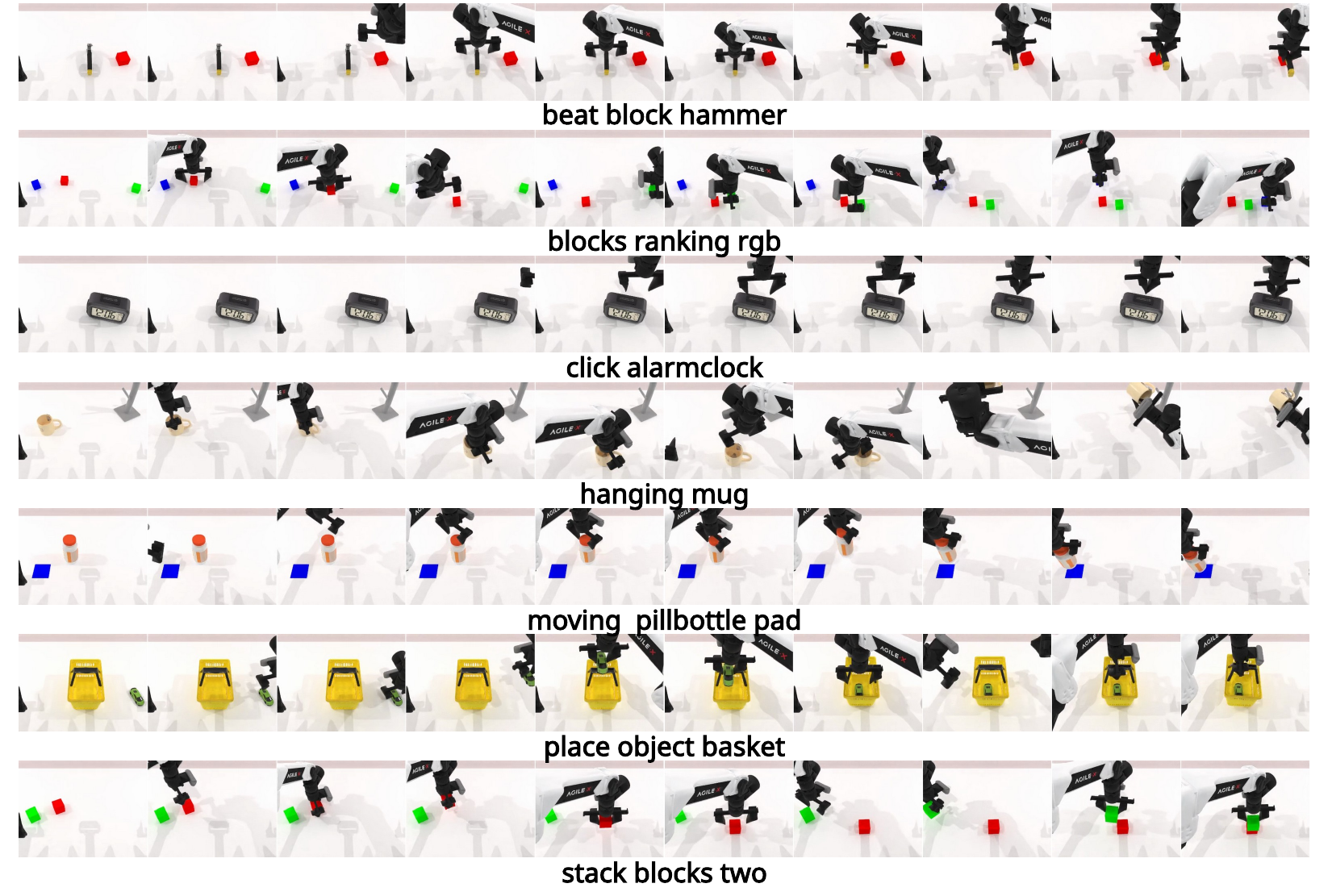}
\caption{Visualizations of FocusVLA executing tasks in the RoboTwin benchmark.}
\label{fig:robotwin_vis}
\vspace{-2mm}
\end{figure}

Fig.~\ref{fig:robotwin_vis} shows representative task executions on the RoboTwin benchmark. We select seven representative manipulation tasks, including \textit{beat}, \textit{ranking}, \textit{click}, \textit{hanging}, \textit{moving}, \textit{place}, and \textit{stack}. These tasks cover diverse types of robotic operations and difficulty levels. As illustrated in the figure, our method achieves strong performance across different task types and difficulty settings.

\begin{figure}[t]
\vspace{-2mm}
\centering
\includegraphics[width=\linewidth]{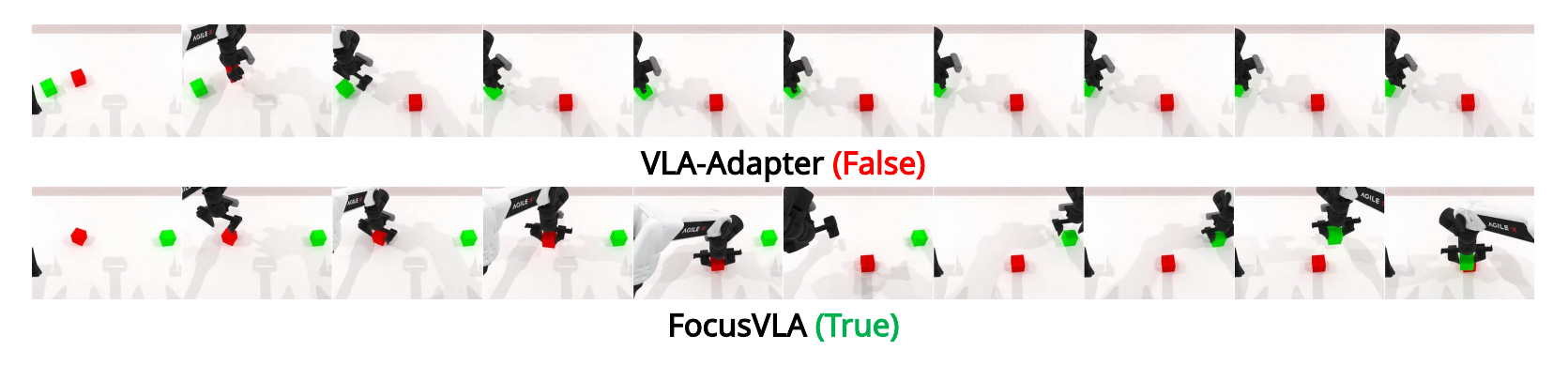}
\caption{Comparison between FocusVLA and VLA-Adapter on the same RoboTwin task.}
\label{fig:robotwin_com}
\vspace{-2mm}
\end{figure}

Fig.~\ref{fig:robotwin_com} compares rollouts of our method and VLA-Adapter on the same task. Our method successfully grasps and stacks the two blocks, while VLA-Adapter fails to grasp the green block and thus cannot complete the stacking task. This result highlights the advantage of our approach in leveraging visual features for precise object manipulation.

\begin{figure}[t] 
\vspace{-2mm}
\centering 
\includegraphics[width=\linewidth]{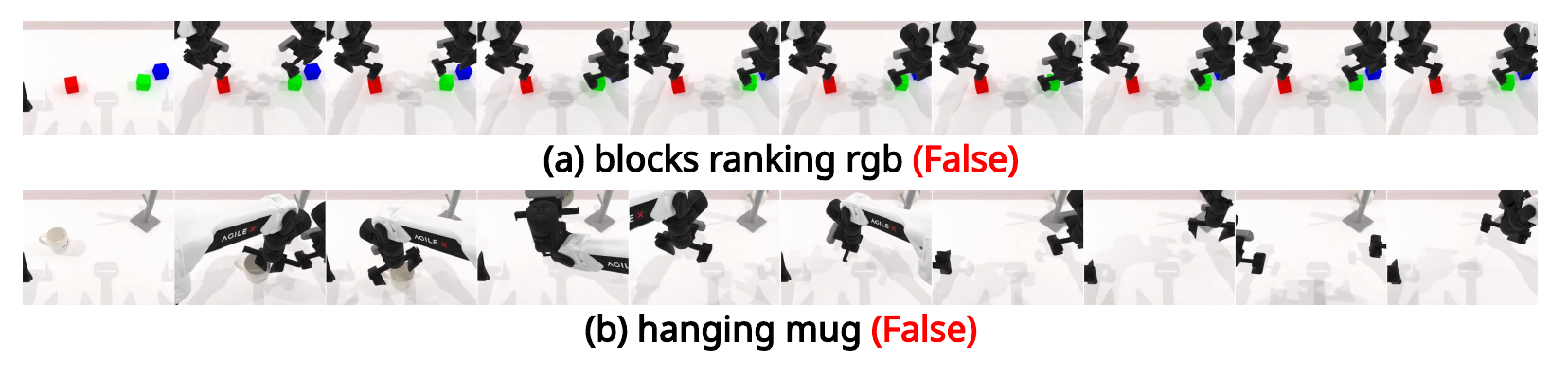} 
\caption{Failure cases on the RoboTwin benchmark.} 
\label{fig:robotwin_failure} 
\vspace{-2mm}
\end{figure}

Fig.~\ref{fig:robotwin_failure} presents two additional failure cases. In Fig.~\ref{fig:robotwin_failure}(a), the failure is caused by coordination issues between the two robot arms. The dual-arm setup increases the possibility of operational conflicts, especially in situations like the one shown in the figure where both arms approach the target object simultaneously. When the action chunk size is small, the robot may alternately or simultaneously operate both arms, which further increases the likelihood of errors. In our experiments, we found that increasing the action chunk size can effectively alleviate this issue. (b) shows a failure case caused by an operational mistake, where the target object is pushed too far away from the table surface, making it unreachable for the robot to grasp again.

\subsection{Visualizations of Real-world Tasks}

\begin{figure}[h]
\vspace{-2mm}
\centering
\includegraphics[width=\linewidth]{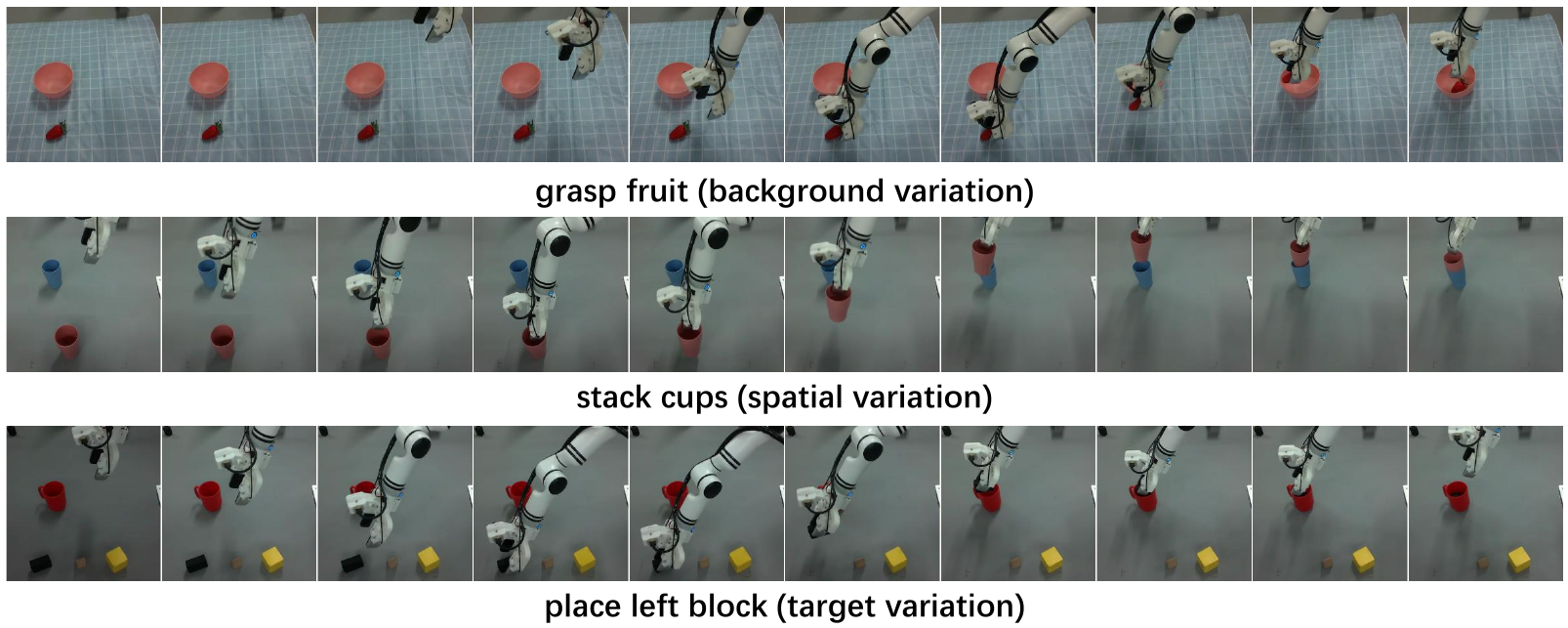}
\caption{Visualizations of FocusVLA executing tasks in the real-world experiments.}
\label{fig:real_vis}
\vspace{-2mm}
\end{figure}

Fig.~\ref{fig:real_vis} presents qualitative examples of FocusVLA executing tasks in real-world environments. We include representative tasks with different types of variations, including \textit{grasp fruit} with background variation, \textit{stack cups} with spatial variation, and \textit{place left block} with target variation. These examples demonstrate that our method can successfully complete manipulation tasks under diverse real-world conditions, even when the scene appearance, object arrangement, or target object changes, suggesting that our method can better leverage visual information to handle diverse variations in real-world environments.

\begin{figure}[h]
\vspace{-2mm}
\centering
\includegraphics[width=\linewidth]{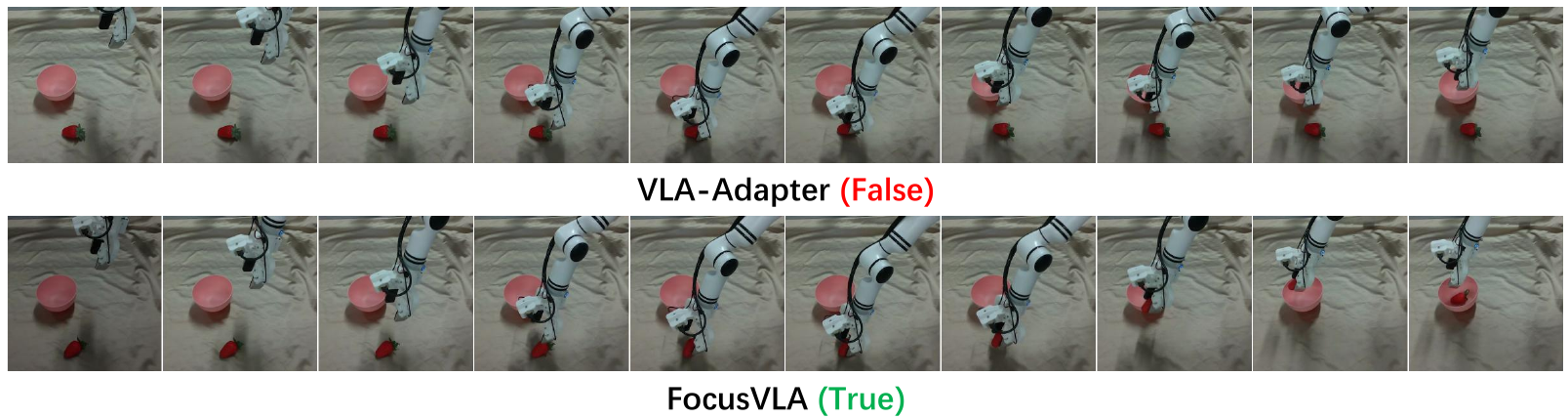}
\caption{Comparison between FocusVLA and VLA-Adapter on the same real-world task.}
\label{fig:real_com}
\vspace{-2mm}
\end{figure}

Fig.~\ref{fig:real_com} compares rollouts of our method and VLA-Adapter on the same real-world task. Our method successfully completes the manipulation, while VLA-Adapter fails. From the rollout, FocusVLA exhibits more accurate target localization and more precise interaction with the object, whereas VLA-Adapter is less reliable in identifying and manipulating the correct target under real-world visual disturbances. This comparison further highlights the advantage of our approach in grounding actions on detailed visual features, which is especially important in real-world environments where background and appearance variations are more complex than in simulation.

\begin{figure}[h]
\vspace{-2mm}
\centering
\includegraphics[width=\linewidth]{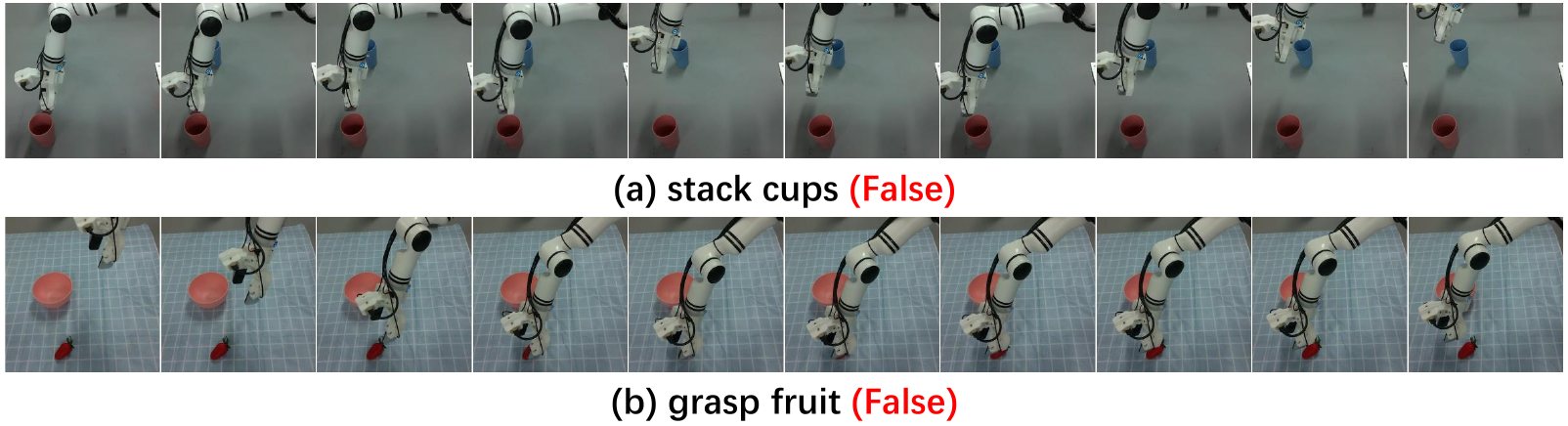}
\caption{Failure cases on the real-world experiments.}
\label{fig:real_failure}
\vspace{-2mm}
\end{figure}

Fig.~\ref{fig:real_failure} shows two representative failure cases of generalization in real-world experiments. In Fig.~\ref{fig:real_failure}(a), the model fails to complete the stacking task because we manually initialize the robot arm between the two cups instead of placing it at the standard initial state. This variation in the initial state causes the model to deviate from the expected trajectory and fail to finish the task successfully. In Fig.~\ref{fig:real_failure}(b), the failure is caused by an environmental generalization issue. We observe that when both the background color and texture change simultaneously, both our method and the baseline suffer a substantial performance drop. This result indicates that the robustness of current VLA models to visual variations remains limited. These examples indicate that real-world manipulation remains challenging, and improving robustness to initialization shifts and environmental variations is an important direction for future work.

\end{document}